\documentclass{mlkd2024cls}
\usepackage{graphicx,amssymb,amsmath,hyperref}
\usepackage{listings}
\usepackage{xcolor}
\usepackage{caption}
\usepackage{float}

\def\QED{\ensuremath{{\square}}}
\def\markatright#1{\leavevmode\unskip\nobreak\quad\hspace*{\fill}{#1}}

\title{Optimizing Data Curation through Spectral Analysis and Joint Batch Selection (SALN)}

\author{Mohammadreza Sharifi\thanks{Department of Computer Science,
        Ferdowsi University of Mashhad, {\tt sharifi.mohammadreza@mail.um.ac.ir}}
}
 
\FirstAuthor{Sharifi}
\SubTitle{ SALN: Spectral-Based Joint Batch Selection}

\begin{document}
\maketitle\thispagestyle{empty}

\begin{abstract}

  In modern deep learning models,
   long training times and large datasets present significant challenges
   to both efficiency and scalability. Effective data curation and
   sample selection are crucial for optimizing the training process
   of deep neural networks. This paper introduces SALN, a method designed to
   prioritize and select samples within each batch rather than from the entire dataset.
   By utilizing jointly selected batches, SALN enhances training efficiency
   compared to independent batch selection. The proposed method applies a
   spectral analysis-based heuristic to identify the most informative data points within each batch,
   improving both training speed and accuracy.
  The SALN algorithm significantly
   reduces training time and enhances accuracy when compared to traditional
   batch prioritization or standard training procedures. It demonstrates up to an 8x reduction in training time and up to a 5\% increase in accuracy over standard training methods. Moreover, SALN achieves better performance and shorter training times compared to Google's JEST method developed by DeepMind. 
The code and Jupyter notebooks are available at \url{github.com/rezasharifi82/SALN}.

\end{abstract}
\vspace{-10pt}
\paragraph{Keywords:} Spectral Analysis, Data Curation, Joint Batch Selection

\section{Introduction}
The effectiveness of a deep learning model's training process largely depends on the quality of the data to which the model is exposed. High-quality data enables the model to learn accurately and generalize well to new, unseen data, improving its overall performance.\cite{budach2022effectsdataqualitymachine}.
Training a deep neural network model on a curated dataset can significantly improve both accuracy and efficiency \cite{JEST}.

There are many well-known methods that apply a kind of data point selection heuristic on the individual manner\cite{coleman2019selectionproxy}.
However, batches of data can have interdependencies \cite{ash2021batch_active_learning} that may lead to much better results than processing them individually \cite{prall2023batch_processing}.
Moreover, there are some approaches for selecting well-informed batches of data and using them in the training process.
These methods represent algorithms or criteria for selecting batches of data that may gain the most information from our dataset \cite{JEST}.

In this paper, a heuristic is proposed to select the most informative batch of data, which can lead to better accuracy as well as shorter training time.
This method is based on the principles of eigenvalues and eigenvectors, leveraging these mathematical tools to identify the core structure and the most informative parts of the data \cite{jolliffe2002principal,luxburg2007tutorial}, rather than using random selection improved by loss control \cite{JEST}.

\section{Related Work}
\textbf{JEST}:
Selecting an informative batch of data can lead to a dramatic improvement in training time and computational resource usage of the model.
Recent research conducted by Google DeepMind has shown that selecting an informative batch of data can lead the model to significantly reduce iterations and computations \cite{JEST}.
Moreover, it can achieve the same or even better performance compared to using all of the data.

% The main idea behind this paper is to first choose a random, unvisited selection from the batch using a filter ratio,$r$,and then 
% attempt to score that specific batch in comparison to other batches.

% In the process of scoring the jointly batch of data, the model starts with a 
% random selection of data, which in my study, it turns to selection mechanism using 
% the eigen vectors and eigen values in an spectral manner.

\textbf{Spectral approaches}:Past research has explored various methods for enhancing the efficiency of machine learning models, particularly through the use of spectral techniques. Eigenvectors and eigenvalues 
have been central to Principal Component Analysis (PCA) and Spectral Clustering, where they are used to identify the most significant dimensions
in the data. These methods have proven effective in reducing the complexity of data and uncovering latent structures, 
which can improve model performance \cite{jolliffe2002principal,luxburg2007tutorial}.
In the context of data selection, prior work has focused primarily on 
filtering or pruning individual samples based on heuristics such as 
importance sampling or noise reduction \cite{sener2018coreset}. 
Additionally, methods such as Core-set selection have been introduced to identify 
a small but representative subset of data for training large-scale models, 
reducing computational costs while maintaining accuracy as good as possible \cite{bachem2017coreset}. 
These approaches, however, often operate on individual samples and do not take significant advantage of the relationships between batches of data points.
Recent works have also explored spectral learning techniques to guide batch selection. 
Methods like SpectralNet \cite{shaham2018spectralnet} utilize the spectral properties 
of data to group data points into batches that better capture 
the underlying structure of the dataset, resulting in more efficient 
training.
% This study builds on these spectral methods by applying them to 
% batch selection scenario. 
% Instead of selecting individual samples, I leverage the eigen-structure 
% of the data's similarity matrix to identify batches that are most 
% representative of the overall dataset. 
% This method ensures that each batch contains the most informative 
% data points, enhancing the model's learning efficiency and performance
% which also experienced a better result of learning time.

\textbf{Batch Selection}:
Batch selection in machine learning has gained considerable attention in 
recent years \cite{JEST} as a strategy to improve the 
efficiency of training models, 
particularly deep neural networks. 
Unlike methods that focus on individual data point selection, 
batch selection methods aim to identify groups of data that are most 
informative or representative together, 
thereby accelerating the learning process and improving model performance \cite{JEST}.
Early work on mini-batch gradient descent \cite{bottou2010large} demonstrated that 
training on small batches of data could significantly reduce computational time 
while retaining much of the accuracy of full-batch methods. 
Since then, several approaches have been developed to 
enhance the effectiveness of batch selection. 
For instance, Core-set selection \cite{sener2018coreset} selects small 
but representative subsets of data that approximate the original dataset's distribution, 
leading to faster convergence and lower resource usage.

\textbf{Active Learning}:
In the context of active learning, batch active learning approaches such as 
BatchBALD \cite{kirsch2019batchbald} 
seek to select batches of data that maximize the information gain during 
training. This contrasts with traditional active learning techniques 
that focus on one data point at a time, demonstrating the benefits of jointly selecting data in batches for training efficiency.

% But now, the problem doesn't involve a deterministic and well-known kind of intention on selecting the data. we need a kind of heuristic on
% selecting the data which can lead us to a better performance. 

\section{Methods}

\subsection{Data Preparation}
Data preprocessing plays a crucial role in the training process 
of machine learning models \cite{guyon2006feature}. 
Proper preprocessing can mitigate issues such as 
overfitting \cite{krizhevsky2012imagenet} and ensure the model generalizes better to unseen data \cite{bishop2006pattern}.
In this study, I have used the Oxford-IIIT Pet Dataset \cite{parkhi2012cats} as the primary dataset and the CIFAR-10 as the secondary dataset for training the model. The Oxford-IIIT Pet Dataset contains images of 37 different breeds of cats and dogs, which are labeled and categorized. 
Meanwhile, the CIFAR-10 dataset consists of 60,000 32x32 color images in 10 classes, with 6,000 images per class and training and test sets of 50,000 and 10,000 images, respectively.

The data augmentation techniques used in this study are as follows:
\begin{itemize}
  \item \textbf{Horizontal Flip}: Random horizontal flipping helps the model improve its robustness to horizontal orientation variations \cite{krizhevsky2012imagenet}.
  \item \textbf{Rotation}: Random rotation by $\alpha =15$ can significantly help the model become invariant to rotational changes \cite{shorten2019survey}.
  \item \textbf{Change on jitter}: To prevent overfitting, one of the most significant augmentations is to adjust the jitter parameters \cite{howard2013some}. That involves the following configurations:
  \begin{itemize}
    \item Brightness: $0.2$
    \item Contrast: $0.2$
    \item Saturation: $0.2$
    \item Hue: $0.1$
  \end{itemize}
  \item \textbf{Resize}: This ensures a uniform input size for all images, which is essential for deep neural networks \cite{simonyan2014very}.
  \item \textbf{Normalization}: Normalizing the pixel values to standardize the input data. The parameters of mean and standard deviation has been set to an appropriate number due to ImageNet statistics. \cite{ioffe2015batch,krizhevsky2012imagenet}.
  \begin{itemize}
    \item \textit{Mean}: $[0.485,0.456,0.406]$
    \item \textit{Std.}: $[0.229,0.224,0.225]$
  \end{itemize}
\end{itemize}

These augmentations were applied to the training data. However, for the test data, since data augmentation is not appropriate \cite{goodfellow2016deep,shorten2019survey} , I have just used resizing, cropping, and normalization.
Also, I split my dataset into training, validation, and test sets.

\subsection{The model}
In this study, I used the pre-trained ResNet-18 model as my main classifier for several reasons:
\begin{itemize}
  \item \textbf{Vanishing Gradients problem}: The residual connections in ResNet-18 efficiently prevent the vanishing gradient problem by allowing information to bypass layers, facilitating the training of deeper networks \cite{he2016resnet}.
  \item \textbf{Strong Transfer Learning Performance}: ResNet-18 is strong enough to capture complex features while still being lightweight, which is enhanced by its pre-trained configuration \cite{kolesnikov2020bit,he2016resnet}.
  \item \textbf{Accessability}: Resnet-18 is available in PyTorch and there is no need to be built from scratch. 
\end{itemize}
So, because of all the reasons listed above, I decided to use the pre-trained version of ResNet-18 as a lightweight, reliable model that is generalized enough to perform significant classification tasks.
As the final layer of this model, I modified the output of the fully-connected layer to have appropriate units, which is corresponds to the number of classes in each dataset.
% \subsection{Early stopping}
% To prevent overfitting, I employed the early stopping procedure, which helps to avoid the 
% model from encountering this issue \cite{goodfellow2016deep}. My model has a patience 
% of 5 in 25 epochs for all experiments which are going to be discussed. I have used this criteria
% on validation loss rather than accuracy which seems to bring a better reasoning to the model \cite{goodfellow2016deep}. 

\subsection{Loss function and Optimizer}
For all of the experiments, I have used \textit{CrossEntropyLoss} as the ideal loss function for
multi-class classification tasks. It combines \textit{LogSoftMax} and negative log likelihood into one function \cite{goodfellow2016deep}.

Meanwhile, for the optimization task, I have used \textit{Stochastic Gradient Descent (SGD)} with $momentum = 0.9$ and $learning \; rate = 0.001$.

\subsection{Joint example selection}
Joint example selection with sigmoid loss is a new method introduced by Google DeepMind \cite{JEST}. It is an intuitive procedure for selecting data in a meaningful way from a batch, rather than relying on random selection, to feed into the model.
This algorithm is based on several important principles \cite{JEST}:
\begin{itemize}
  \item \textbf{Loss calculation}: This algorithm needs both reference loss and learner loss.
  \begin{itemize}
    \item \textit{Reference loss}: The model's current performance on each example.
    \item \textit{Learner loss}: A reference loss that could represent a baseline(performance of a reference model).
  \end{itemize}
  \item \textbf{Filter ratio}: This ratio is a hyperparameter which is used to select the amount of data.
  \item \textbf{Interaction}: This algorithm uses an intuitive procedure to prioritize those data from a batch which are most informative rather than others. In fact, it 
  calculates the impression of selected data on the remaining ones, as well as the remaining ones on the selected data. In the other word, it can measure how the selected data could interact with a remaining one.
  \item \textbf{Not using old ones}: This algorithm uses a probability distribution which assigns a very-low weight to the previously selected samples. This action would guarantee the algorithm will not choose the previously-selected samples. 
  \item \textbf{Size of chunks}: This algorithm has a parameter that adjusts the number of data points representable in each chunk, associated with the filter ratio.
\end{itemize}

\subsection{SALN(My method)}
In this approach, I've utilized spectral analysis to identify and prioritize the most informative data based on their structural significance within a batch.
 Since using spectral approaches can leverage the core structure of the data \cite{luxburg2007tutorial,ng2002spectral}, my methodology consists of the following steps:
\begin{enumerate}
  \item \textbf{Feature extraction}: Using a pre-trained model without fine-tuning to extract features from the whole dataset can dramatically reduce time and computational resources \cite{goodfellow2016deep}.
  This step is crucial in this study in order to have a comprehensive features for each data point. I have used ResNet-50
  as a reference pre-trained model to extract the features of each data and convert them to a vector.
  \item \textbf{Compute similarity matrix}: For each batch of data,computing a similarity matrix $S$ using \textit{cosine similarity} can significantly improve the algorithm's understanding of meaningful structures, which is useful for further spectral analysis. \cite{singhal2001information,luxburg2007tutorial}.
  \item \textbf{Degree matrix}: The degree matrix $D$ is a diagonal 
  matrix where each element $D_{ii}$ represents the sum 
  of the similarities of all relevant 
  data point $i$ in a dataset. 
  The degree matrix plays a key role in defining the Laplacian matrix \cite{chung1997spectral}.
  \item \textbf{Laplacian matrix}: The Laplacian matrix $L$ is derived from the degree matrix $D$
   and the similarity matrix $S$, typically computed as $L=D-S$. The Laplacian matrix
   encodes the structure of the dataset by capturing the relationships between data points.
  \item \textbf{Eigen decomposition}: To score the data and their relationships, the second smallest eigenvalue of the Laplacian matrix is calculated, which corresponds to the Fiedler value \cite{fiedler1973algebraic}.
   This approach utilizes the spectral properties of the Laplacian to identify samples that are essential for preserving the structural integrity of the current batch \cite{fiedler1973algebraic,luxburg2007tutorial,ng2002spectral}.
  \item \textbf{Informative indices}: The resulting vector consists of indices corresponding to the most informative data points in the current batch. The size of this vector is determined by the filter ratio. 
  \item \textbf{Joint batch sampling}: Similar to the \textit{JointExample Selection} algorithm \cite{JEST}, the introduced method implicitly uses the similar criteria, with a slight difference in the scoring mechanism, which is handled through spectral analysis.
\end{enumerate}

\subsection{Algorithm}
The following algorithm describes the procedure for \textit{Joint batch sampling} using Laplacian matrix and spectral analysis.
 The algorithm takes as input a batch of feature vectors corresponding to
 current batch's data. These vectors had been extracted in the manner which has described in previous section.
  Now the algorithm selects data points which
 are the most informative in this batch relative to filter ratio parameter.
The \textit{filter ratio} parameter, is a hyper-parameter which can specify how many data should be selected from this batch.
 At the end, the algorithm returns the most informative indices, corresponding to extracted properties from Fiedler vector.
 These indices are going to be used further in the training loop with an specific intention to reduce number of processing input data.
 
 \begin{lstlisting}[language=Python, caption={Joint batch sampling using spectral analysis(SALN)}]
  def Joint_batch_sampling(current_batch_data, filter_ratio=0.8):
          
      # Number of images (samples) in this batch
      n_images = current_batch_data.size(0)
      
      batch_data = current_batch_data.detach().cpu()
      
      # Flatten the batch data to 2D
      batch_data_flat = batch_data.view(n_images, -1).numpy()
      
      # Number of samples to select
      n_draws = int(n_images * (1 - filter_ratio))
  
      # Compute cosine similarity matrix
      similarity_matrix = cosine_similarity(batch_data_flat)
  
      # Compute the degree matrix
      degree_matrix = np.diag(similarity_matrix.sum(axis=1))
  
      # Compute the Laplacian matrix
      laplacian_matrix = degree_matrix - similarity_matrix
  
      # Find eigens of Laplacian matrix
      eigenvalues, eigenvectors = np.linalg.eigh(laplacian_matrix)
  
      # Extract the Fiedler vector
      fiedler_vector = eigenvectors[:, 1]
  
      # Sort the absolute values of the Fiedler
      informative_indices = np.argsort(np.abs(fiedler_vector))[-n_draws:]
  
      return informative_indices
  \end{lstlisting}

\newpage

\section{Experiments}
All the experiments were conducted on those datasets which have described in the previous section.
 The experiments were conducted on a Google-Colab service wit T4-GPU. Number of epochs and other important factors were kept the same.
  The experiments have been conducted on the same model, ResNet-18, with the same hyper-parameters and configurations.
  The only difference between the experiments is the method of selecting the data points in each batch during the 25 epochs.

\subsection{Primary Dataset: Oxford-IIIT Pet }
The primary dataset used in this study is the Oxford-IIIT Pet Dataset, which contains images of 37 different breeds of cats and dogs. The dataset is labeled and categorized, making it suitable for classification tasks.
  \subsubsection{Standard training and SALN}
The following experiment has been conducted to compare the performance of
 the standard training method as the baseline, with the proposed SALN method.
 The standard training method uses no specific selection criteria in order to select the data,
  while SALN employs spectral analysis to identify the most informative data points in each batch.
  The standard training method processes the entire dataset without
   any form of selective sampling or data reduction.
    Every data point is used during training,
    and no prioritization or filtering is applied to optimize the learning process.
  The results of this experiment consisted from several reports:
  \begin{itemize}
    \item Comparison between training and validation accuracy(\%) of SALN and standard
     training, which has represented in Table~\ref{tab:accuracy_comparison}.
    \begin{table}[H]
        \centering
        \begin{tabular}{|l|c|c|}
            \hline
            \textbf{Method} & \textbf{Training} & \textbf{Validation} \\ 
            \hline
            SALN   & 96.5\%  & 94.02\%  \\  
            \hline
            Standard Training & 86.1\%  & 76.49\%  \\ 
            \hline
        \end{tabular}
        \caption{Accuracy Comparison of SALN and Standard Training}
    
        \label{tab:accuracy_comparison}
    \end{table}

    \item Comparison between training and validation loss of SALN and standard
     training, which has represented in Table~\ref{tab:loss_comparison}.
    
    \begin{table}[H]
      \centering
      \begin{tabular}{|l|c|c|}
          \hline
          \textbf{Method} & \textbf{Training} & \textbf{Validation} \\ 
          \hline
          SALN   & 0.1606  & 0.2102  \\  
          \hline
          Standard Training & 0.4983 & 0.6546  \\ 
          \hline
      \end{tabular}
      \caption{Loss Comparison of SALN and Standard Training}
  
      \label{tab:loss_comparison}
  \end{table}

    \item Comparison between test set accuracy of SALN and standard training,
     which has represented in Table~\ref{tab:test_comparison}.
    \begin{table}[H]
      \centering
      \begin{tabular}{|l|c|}
          \hline
          \textbf{Method} & \textbf{Test Accuracy} \\ 
          \hline
          SALN   & 86.8\%   \\  
          \hline
          Standard Training & 82.02\%  \\ 
          \hline
      \end{tabular}
      \caption{Comparison of SALN and Standard Training Test results}
  
      \label{tab:test_comparison}
  \end{table}

  \item Comparison between training time of SALN and standard training,
     which has represented in Table~\ref{tab:time_comparison}.
    \begin{table}[H]
      \centering
      \begin{tabular}{|l|c|}
          \hline
          \textbf{Method} & \textbf{Training Time (minute)} \\ 
          \hline
          SALN   & 6.31   \\  
          \hline
          Standard Training & 24.48  \\ 
          \hline
      \end{tabular}
      \caption{Training time Comparison of SALN and Standard Training}
  
      \label{tab:time_comparison}
  \end{table}
  
  \item Standard Training accuracy and loss curves over 25 epochs,
   which has represented in Figure~\ref{fig:Standard_curve}.
   \begin{figure}[H]
    \centering
    \includegraphics[width=0.99\columnwidth]{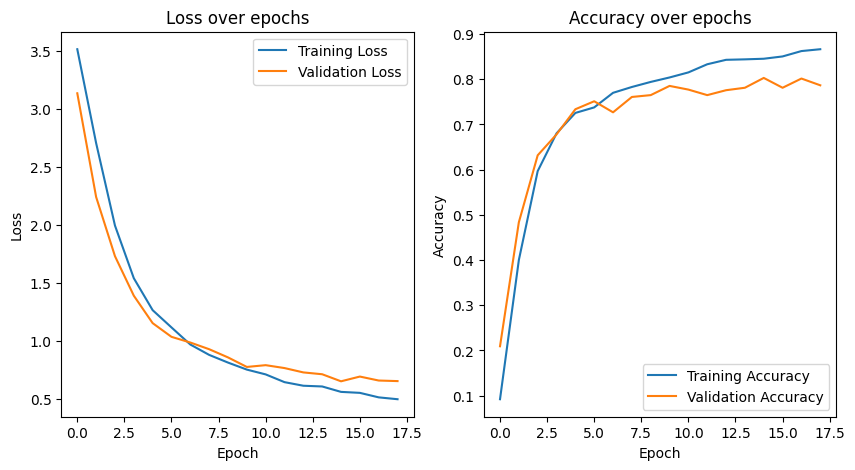}  % Insert loss curve image
    \caption{Accuracy and Loss Curves of \textbf{Standard-Training} Method over epochs}
    \label{fig:Standard_curve}
\end{figure}

\item SALN accuracy and loss curves over 25 epochs,
 which has represented in Figure~\ref{fig:SALN_curve}.

\begin{figure}[H]
    \centering
    \includegraphics[width=1\columnwidth]{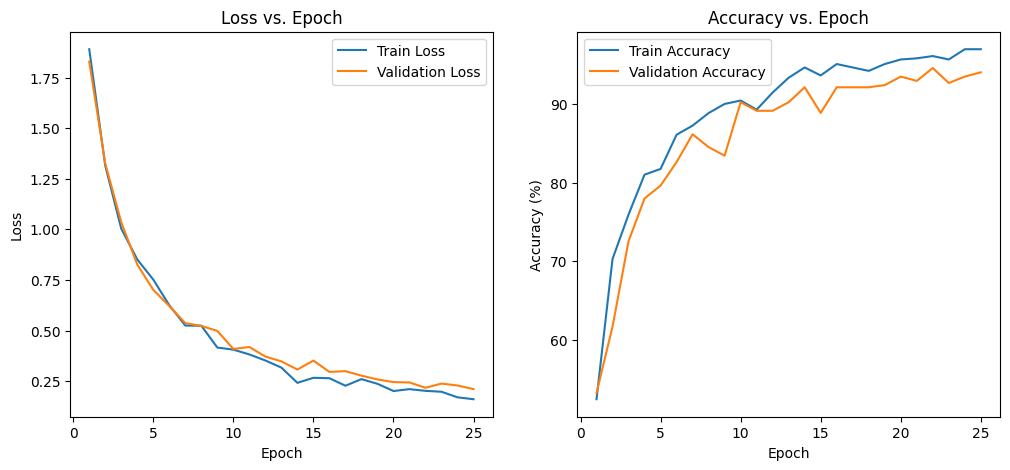}  % Insert accuracy curve image
    \caption{Accuracy and Loss Curves of \textbf{SALN} Method over epochs}
    \label{fig:SALN_curve}
\end{figure}
  \end{itemize}

  %----------------------------------------------------------------------------------------------------------------------------
  \subsubsection{JEST and SALN}
  The following experiment has been conducted to compare the performance of
   the JEST method which has proposed by Google DeepMind, and the SALN method.
  The JEST method uses a specific selection criteria in order to select the informative data.
  This criteria were discussed in the previous section.
   The results of this experiment consisted from several reports:

   %  The standard training method uses no specific selection criteria in order to select the data,
  %   while SALN employs spectral analysis to identify the most informative data points in each batch.
  %   The standard training method processes the entire dataset without
  %    any form of selective sampling or data reduction.
  %     Every data point is used during training,
  %     and no prioritization or filtering is applied to optimize the learning process.
  %   The results of this experiment consisted from several reports:
    \begin{itemize}
      \item Comparison between training and validation accuracy(\%) of SALN and JEST,
       which has represented in Table~\ref{tab:SJ_accuracy_comparison}.
      \begin{table}[h]
          \centering 
          \begin{tabular}{|l|c|c|}
              \hline
              \textbf{Method} & \textbf{Training} & \textbf{Validation} \\ 
              \hline
              SALN   & 96.5\%  & 94.02\%  \\  
              \hline
              JEST & 79.59\%  & 75.54\%  \\ 
              \hline
          \end{tabular}
          \caption{Accuracy Comparison of SALN and JEST}
      
          \label{tab:SJ_accuracy_comparison}
      \end{table} 
  
      \item Comparison between training and validation loss of SALN and JEST,
       which has represented in Table~\ref{tab:SJ_loss_comparison}.
      
      \begin{table}[h]
        \centering
        \begin{tabular}{|l|c|c|}
            \hline
            \textbf{Method} & \textbf{Training} & \textbf{Validation} \\ 
            \hline
            SALN   & 0.1606  & 0.2102  \\  
            \hline
            JEST & 0.7309 & 0.8060  \\ 
            \hline
        \end{tabular}
        \caption{Loss Comparison of SALN and JEST}
    
        \label{tab:SJ_loss_comparison}
    \end{table}

      \item Comparison between test set accuracy of SALN and JEST,
       which has represented in Table~\ref{tab:SJ_test_comparison}.
      \begin{table}[H]
        \centering
        \begin{tabular}{|l|c|}
            \hline
            \textbf{Method} & \textbf{Test Accuracy} \\ 
            \hline
            SALN   & 86.8\%   \\  
            \hline
            JEST & 87.55\%  \\ 
            \hline
        \end{tabular}
        \caption{Loss Comparison of SALN and JEST}
    
        \label{tab:SJ_test_comparison}
    \end{table}
  
    \item Comparison between training time of SALN and JEST,
       which has represented in Table~\ref{tab:SJ_time_comparison}.
      \begin{table}[H]
        \centering
        \begin{tabular}{|l|c|}
            \hline
            \textbf{Method} & \textbf{Training Time (minute)} \\ 
            \hline
            SALN   & 6.31   \\  
            \hline
            JEST & 14.87  \\ 
            \hline
        \end{tabular}
        \caption{Training time Comparison of SALN and JEST}
    
        \label{tab:SJ_time_comparison}
    \end{table}
    
    \item JEST accuracy and loss curves over 25 epochs,
     which has represented in Figure~\ref{fig:JEST_curve}.
     \begin{figure}[H]
      \centering
      \includegraphics[width=0.99\columnwidth]{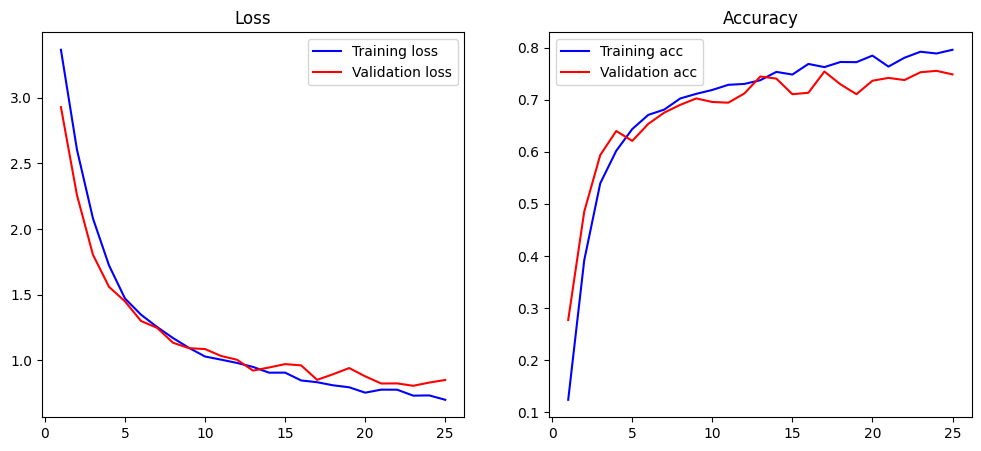}  % Insert loss curve image
      \caption{Accuracy and Loss Curves of \textbf{JEST} Method over epochs}
      \label{fig:JEST_curve}
  \end{figure}
  
  \item SALN accuracy and loss curves over 25 epochs,
   which has represented in Figure~\ref{fig:SALN_curve}.
    \end{itemize}

    \subsection{Secondary Dataset: CIFAR-10 }
    The secondary dataset used in this study is the CIFAR-10 dataset, which contains 60,000 32x32 color images in 10 classes, with 6,000 images per class and training and test sets of 50,000 and 10,000 images, respectively.

    \subsubsection{Standard training and SALN}
    Same as the primary dataset, the following experiment has been conducted to compare the performance of
      the standard training method as the baseline, with the proposed SALN method.
      The results of this experiment consisted from several reports:
      \begin{itemize}
        \item Comparison between training and validation accuracy(\%) of SALN and standard
         training, which has represented in Table~\ref{tab:accuracy_comparison_cifar}.
        \begin{table}[H]
            \centering
            \begin{tabular}{|l|c|c|}
                \hline
                \textbf{Method} & \textbf{Training} & \textbf{Validation} \\ 
                \hline
                SALN   & 81.99\%  & 84.08\%  \\  
                \hline
                Standard Training & 85.54\%  & 86.82\%  \\ 
                \hline
            \end{tabular}
            \caption{Accuracy Comparison of SALN and Standard Training}
        
            \label{tab:accuracy_comparison_cifar}
        \end{table}
    
        \item Comparison between training and validation loss of SALN and standard
         training, which has represented in Table~\ref{tab:loss_comparison_cifar}.
        
        \begin{table}[H]
          \centering
          \begin{tabular}{|l|c|c|}
              \hline
              \textbf{Method} & \textbf{Training} & \textbf{Validation} \\ 
              \hline
              SALN   & 0.5080  & 0.4607  \\  
              \hline
              Standard Training & 0.4055 & 0.3664  \\ 
              \hline
          \end{tabular}
          \caption{Loss Comparison of SALN and Standard Training}
      
          \label{tab:loss_comparison_cifar}
      \end{table}

        \item Comparison between test set accuracy of SALN and standard training,
         which has represented in Table~\ref{tab:test_comparison_cifar}.
        \begin{table}[H]
          \centering
          \begin{tabular}{|l|c|}
              \hline
              \textbf{Method} & \textbf{Test Accuracy} \\ 
              \hline
              SALN   & 82.46\%   \\  
              \hline
              Standard Training & 82.56\%  \\ 
              \hline
          \end{tabular}
          \caption{Comparison of SALN and Standard Training Test results}
      
          \label{tab:test_comparison_cifar}
      \end{table}
    
      \item Comparison between training time of SALN and standard training,
         which has represented in Table~\ref{tab:time_comparison_cifar}.
        \begin{table}[H]
          \centering
          \begin{tabular}{|l|c|}
              \hline
              \textbf{Method} & \textbf{Training Time (minute)} \\ 
              \hline
              SALN   & 22.66   \\  
              \hline
              Standard Training & 39.41  \\ 
              \hline
          \end{tabular}
          \caption{Training time Comparison of SALN and Standard Training}
      
          \label{tab:time_comparison_cifar}
      \end{table}
      
      \item Standard Training accuracy and loss curves over 25 epochs,
       which has represented in Figure~\ref{fig:Standard_curve_cifar}.
       \begin{figure}[H]
        \centering
        \includegraphics[width=0.99\columnwidth]{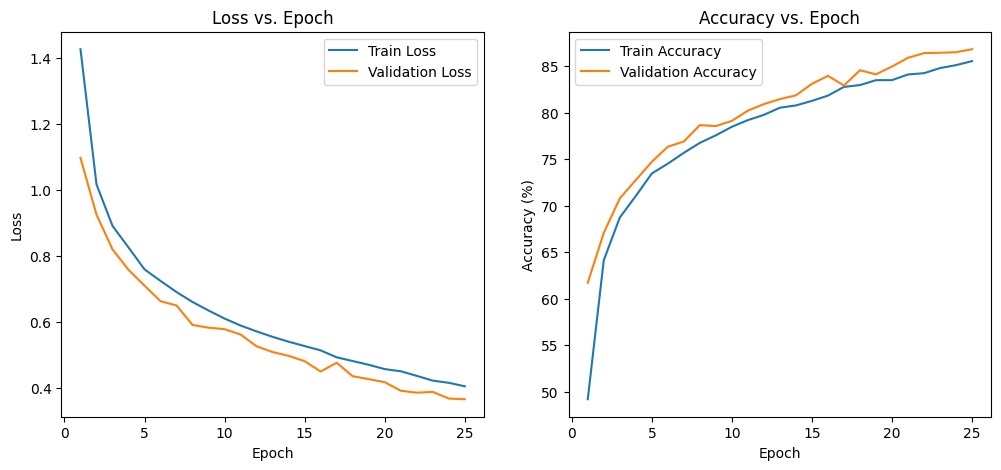}  % Insert loss curve image
        \caption{Accuracy and Loss Curves of \textbf{Standard-Training} Method over epochs}
        \label{fig:Standard_curve_cifar}
    \end{figure}
    
    \item SALN accuracy and loss curves over 25 epochs,
     which has represented in Figure~\ref{fig:SALN_curve_cifar}.

    \begin{figure}[H]
        \centering
        \includegraphics[width=1\columnwidth]{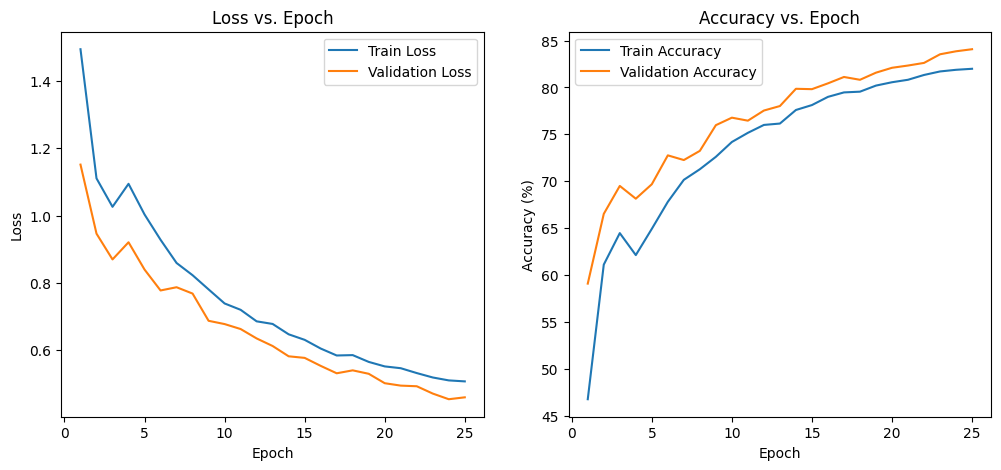}  % Insert accuracy curve image
        \caption{Accuracy and Loss Curves of \textbf{SALN} Method over epochs}
        \label{fig:SALN_curve_cifar} 
    \end{figure}
      \end{itemize}
    
      %----------------------------------------------------------------------------------------------------------------------------
      \subsubsection{JEST and SALN}
      Same as primary dataset, the following experiment has been conducted to compare the performance of the JEST method which has proposed by Google DeepMind, and the SALN method.
       The results of this experiment consisted from several reports:
    
       %  The standard training method uses no specific selection criteria in order to select the data,
      %   while SALN employs spectral analysis to identify the most informative data points in each batch.
      %   The standard training method processes the entire dataset without
      %    any form of selective sampling or data reduction.
      %     Every data point is used during training,
      %     and no prioritization or filtering is applied to optimize the learning process.
      %   The results of this experiment consisted from several reports:
        \begin{itemize}
          \item Comparison between training and validation accuracy(\%) of SALN and JEST,
           which has represented in Table~\ref{tab:SJ_accuracy_comparison_cifar}.
          \begin{table}[h]
              \centering 
              \begin{tabular}{|l|c|c|}
                  \hline
                  \textbf{Method} & \textbf{Training} & \textbf{Validation} \\ 
                  \hline
                  SALN   & 81.99\%  & 84.08\%  \\  
                  \hline
                  JEST & 72.81\%  & 76.86\%  \\ 
                  \hline
              \end{tabular}
              \caption{Accuracy Comparison of SALN and JEST}
          
              \label{tab:SJ_accuracy_comparison_cifar}
          \end{table} 
      
          \item Comparison between training and validation loss of SALN and JEST,
           which has represented in Table~\ref{tab:SJ_loss_comparison_cifar}.
          
          \begin{table}[h]
            \centering
            \begin{tabular}{|l|c|c|}
                \hline
                \textbf{Method} & \textbf{Training} & \textbf{Validation} \\ 
                \hline
                SALN   & 0.5080  & 0.4607  \\  
                \hline
                JEST & 0.7832 & 0.6732  \\ 
                \hline
            \end{tabular}
            \caption{Loss Comparison of SALN and JEST}
        
            \label{tab:SJ_loss_comparison_cifar}
        \end{table}

          \item Comparison between test set accuracy of SALN and JEST,
           which has represented in Table~\ref{tab:SJ_test_comparison_cifar}.
          \begin{table}[H]
            \centering
            \begin{tabular}{|l|c|}
                \hline
                \textbf{Method} & \textbf{Test Accuracy} \\ 
                \hline
                SALN   & 82.46\%   \\  
                \hline
                JEST & 77.72\%  \\ 
                \hline
            \end{tabular}
            \caption{Loss Comparison of SALN and JEST}
        
            \label{tab:SJ_test_comparison_cifar}
        \end{table}
      
        \item Comparison between training time of SALN and JEST,
           which has represented in Table~\ref{tab:SJ_time_comparison}.
          \begin{table}[H]
            \centering
            \begin{tabular}{|l|c|}
                \hline
                \textbf{Method} & \textbf{Training Time (minute)} \\ 
                \hline
                SALN   & 22.66   \\  
                \hline
                JEST & 37.68  \\ 
                \hline
            \end{tabular}
            \caption{Training time Comparison of SALN and JEST}
        
            \label{tab:SJ_time_comparison}
        \end{table}
        
        \item JEST accuracy and loss curves over 25 epochs,
         which has represented in Figure~\ref{fig:JEST_curve_cifar}.
         \begin{figure}[H]
          \centering
          \includegraphics[width=0.99\columnwidth]{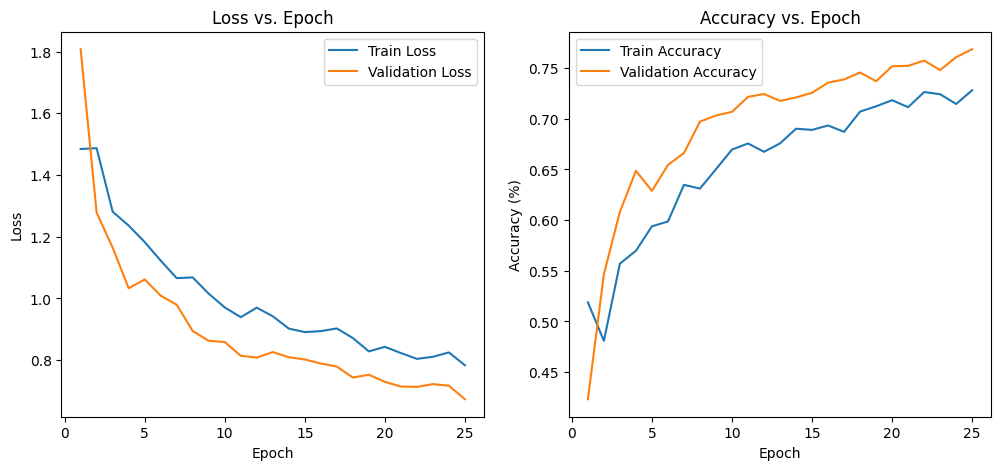}  % Insert loss curve image
          \caption{Accuracy and Loss Curves of \textbf{JEST} Method over epochs}
          \label{fig:JEST_curve_cifar}
      \end{figure}
      
      \item SALN accuracy and loss curves over 25 epochs,
       which has represented in Figure~\ref{fig:SALN_curve_cifar}.
        \end{itemize}

    \subsection{SALN data-selection visualization}
    In this section, the data selection process of the SALN algorithm will be visualized.
    The visualization will show the top 50\% data points that were selected by the algorithm in each batch.
    Here, I have selected Batch No.0 and Batch No.1 from the Oxford-IIIT Pet dataset and,
      Batch No.50 and Batch No.51 from the CIFAR-10 dataset.
    
    \begin{figure}[H]
      \centering
      \includegraphics[width=0.99\columnwidth]{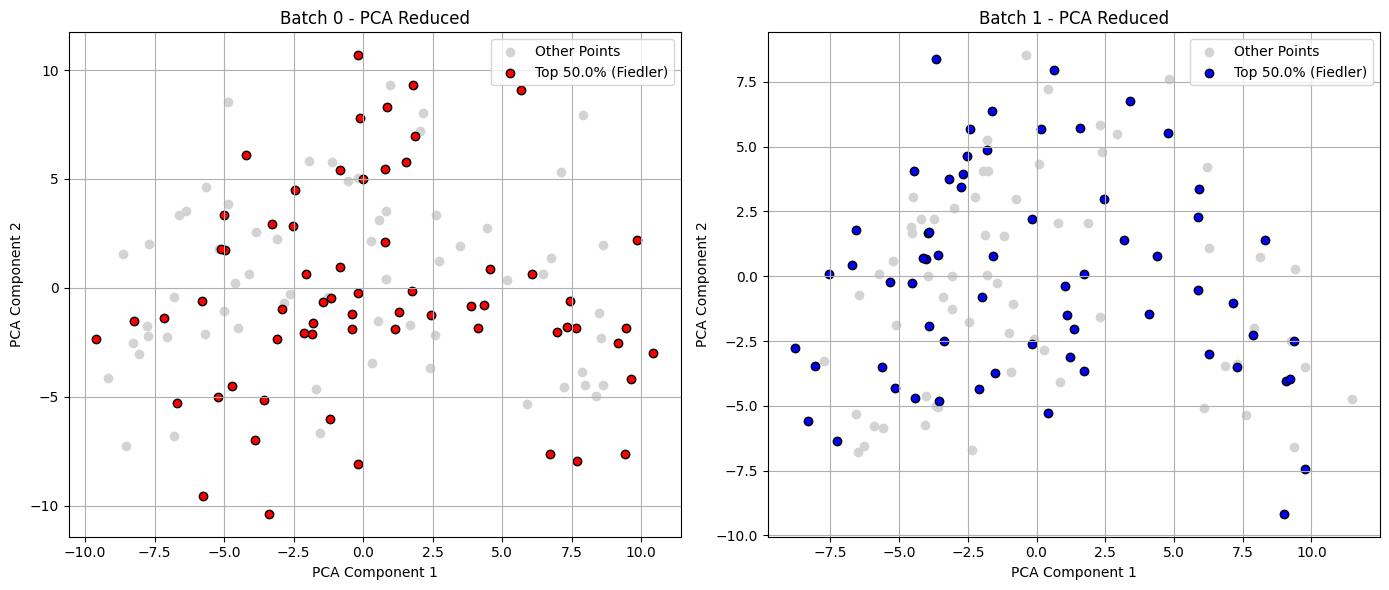}  % Insert loss curve image
      \caption{SALN Data Selection Visualization of Oxford-IIIT Pet Dataset}
      \label{fig:SALN_visualization}
  \end{figure}

  \begin{figure}[H]
    \centering
    \includegraphics[width=0.99\columnwidth]{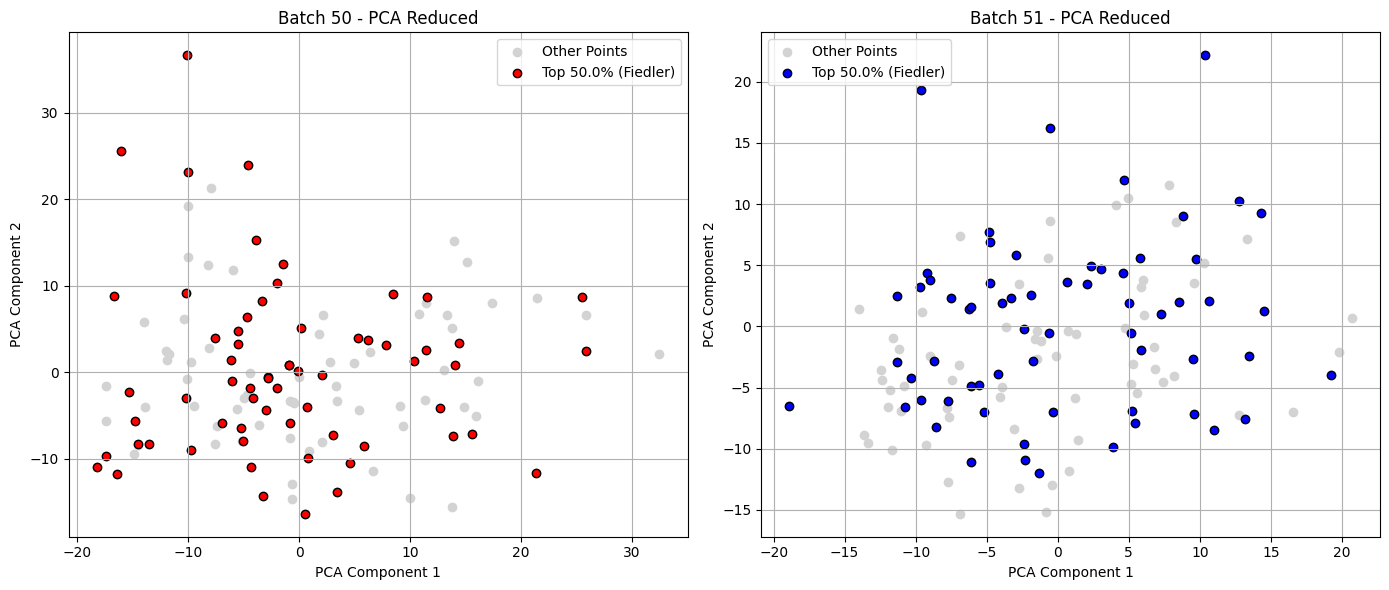}  % Insert loss curve image
    \caption{SALN Data Selection Visualization of CIFAR-10 Dataset}
    \label{fig:SALN_visualization_cifar}
\end{figure}

    \subsection{Analysis of SALN Weights and Filters}
  In this section, an analysis of the model which has trained on Oxford-IIIT Pet dataset 
  will be presented. Since ResNet-18 is well-known, the focus will be on the details of the final fully connected layer.
  The analysis would be in two parts:
  \begin{itemize}
    \item \textbf{Weights Heatmap of fully connected layer}: The heatmap in Figure~\ref{fig:SALN_heat}
     visualizes the weights of the final fully
     connected (fc) layer in the model.
     Each row represents a neuron in the fc
     layer, and each column corresponds to a
     weight connected to an input feature.
     \begin{figure}[h]
      \centering
      \includegraphics[width=0.8\columnwidth]{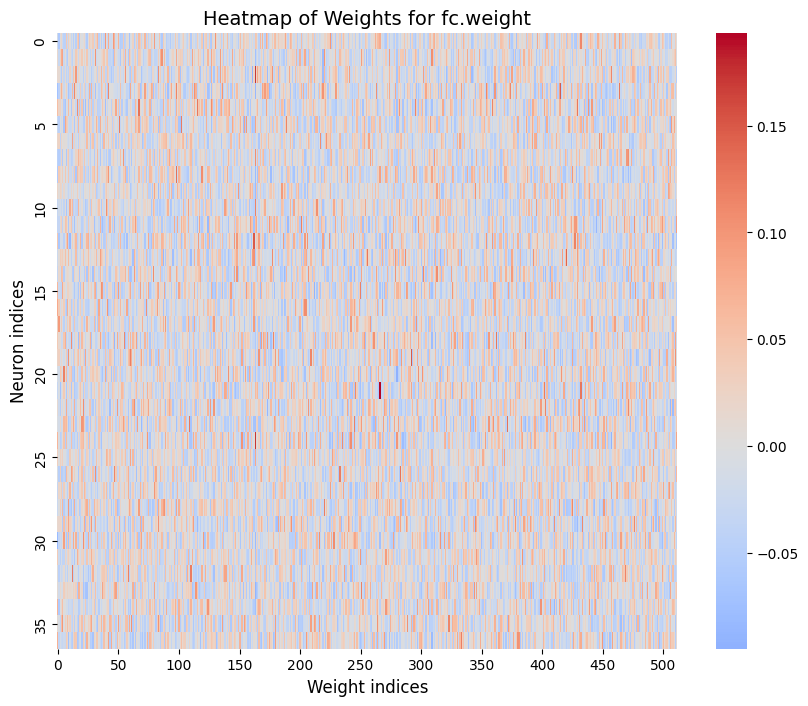}  % Insert loss curve image
      \caption{SALN Weights Heatmap of fully connected layer}
      \label{fig:SALN_heat} 
  \end{figure}

  \item \textbf{Weights Distribution of fully connected layer}:Figure~\ref{fig:SALN_fc}
   shows the distribution of
   the weights in the final
   fully connected (fc) layer.
   The histogram represents how the
   weights are spread across all neurons in this layer. A large concentration of weights near zero suggests that many connections have small magnitudes, which is often seen in models that are well-regularized. The range of the weight values indicates how much emphasis the model places on different input features, with any outliers potentially showing connections that have a stronger influence on the final predictions.

  \begin{figure}[h]
    \centering
    \includegraphics[width=0.8\columnwidth]{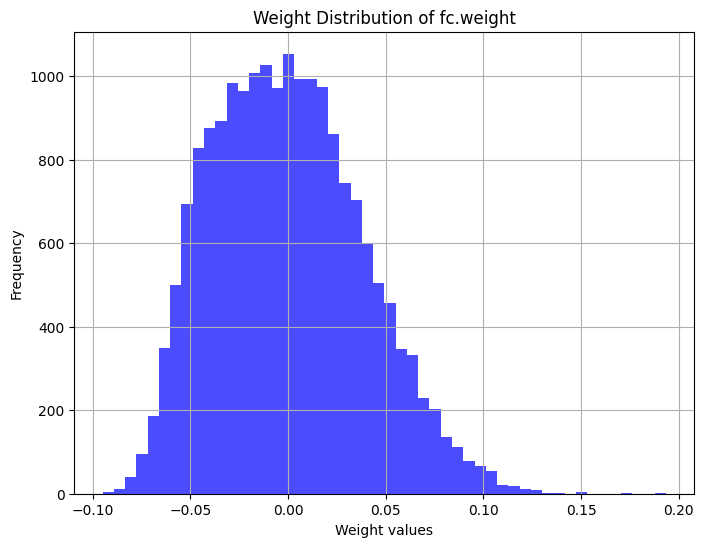}  % Insert loss curve image
    \caption{SALN Weights Heatmap of fully connected layer}
    \label{fig:SALN_fc} 
\end{figure}
 
\end{itemize}

  %----------------------------------------------------------------------------------------------------------------------------

% \section{Submission Link}
% The submission must be submitted through Microsoft’s Conference Management Toolkit; here is the link: \url{https://cmt3.research.microsoft.com/MLKD2024}
\section{Discussion}
The proposed method, SALN, introduces a
 novel approach to jointly batch sampling
 through the application of spectral methodologies.
 This technique enhances the selection of data points
 during training, aiming to accelerate the learning process
 when compared to standard training methods.

The results from our experiments
 indicate that SALN offers a substantial reduction in training time by utilizing an optimized jointly batch sampling mechanism. The spectral techniques employed in this method, enable the model to prioritize the most informative data. And thereby, improving both the speed and effectiveness of training procedure.

Additionally, the experiments demonstrate the high potential of \textit{data bootstrapping} in deep neural network training process. By focusing on batches that maximize the learnability, SALN ensures that the model is consistently exposed to the most challenging and relevant examples, which contributes to more efficient and robust learning.

In summary, SALN provides a promising advancement in batch sampling strategies, offering significant improvements in training efficiency, using less time than JEST, without sacrificing model performance.

\newpage
%---------------------------- Bibliography -------------------------------

% Please add the contents of the .bbl file that you generate,  or add bibitem entries manually if you like.
% The entries should be in alphabetical order
\newpage
\small
\newpage
\bibliographystyle{abbrv}

%---------------------------- Appendix -------------------------------

\end{document}